\documentclass{article}
\usepackage{spconf,amsmath,graphicx}

\title{Contrastive  Graph Multimodal  Model for Text Classification in Videos}

\name{Ye Liu\sthanks{Equal contribution. e-mails: rafelliu@tencent.com. chonglu@tencent.com} \qquad Changchong Lu$^{\textasteriskcentered}$ \qquad Chen Lin \qquad Di Yin \qquad Bo Ren}

\address{Tencent}
      
\begin{document}
%
\maketitle
\begin{abstract}


The  extraction  of text  information in videos  serves as  a  critical  step towards  semantic understanding of videos. It usually involved in two steps: (1) text recognition and (2) text classification. To localize texts in videos, we can resort to large numbers of text recognition methods based on OCR technology. However, to our knowledge,  there is no existing work focused on the second step of video text classification, which will limit the guidance to downstream  tasks such as video indexing and browsing. In this paper, we are the first  to address this new task of video text classification by fusing multimodal information to deal with the challenging scenario where different types of video texts may be confused with various colors, unknown fonts and complex layouts. In addition, we tailor a specific module called CorrelationNet to  reinforce  feature representation by explicitly extracting layout information. Furthermore, contrastive learning is utilized to explore inherent connections  between  samples  using  plentiful  unlabeled  videos. 
Finally, we construct a new well-defined industrial dataset from the news domain, called  TI-News, which is dedicated to building and evaluating video text recognition and classification applications. Extensive  experiments  on  TI-News  demonstrate  the  effectiveness  of our  method.





\end{abstract}
\section{Introduction}
\label{sec:intro}

With  massive  video  data  generated  everyday,  the extraction of textual information from videos is an essential work in many video applications. Texts embedded in videos usually carry rich semantic description about the video content  and this information gives high-level index for a content-based video indexing and browsing. 
The extraction of video text information proceeds in two main steps: text  recognition followed by text  classification.
To localize  texts in  videos, one can resort to  large numbers of text detection methods based on OCR technology. A number of deep neural network (DNN) based methods \cite{liao2018rotation,tian2016detecting,zhou2017east} have been proposed to automatically learn effective features of text and localize text in an image using variant DNN models such as convolutional neural network (CNN) and recurrent neural network (RNN).  To make use of complementary text cues in multiple related video frames, some multi-frame video text detection methods \cite{wang2018scene,yang2017tracking},  such as spatial-temporal analysis, have been proposed to further improve the overall detection performance on the basis of single-image text detection methods. 

In this paper, we concentrate on the second step of video text classification, which plays a more important role in downstream  tasks such as video indexing and browsing. For example, caption summarizes the whole video  
and represents the most important or relevant information within the original video content. Subtitle expresses what the speaker thought, which contains abundant details describing video events.  However, to our knowledge, there is no existing work attempting to address this task, which guides us to turn to some similar tasks, such as text classification and scene/caption classification. Text classification is a fundamental problem behind many research topics in Natural Language Processing (NLP), such as topic categorization, sentiment analysis, relation extraction, etc. Existing text classification methods can be classified into bag-of-words/n-grams models \cite{wang2012baselines}, CNN \cite{le2018convolutional,li2020improving}, RNN \cite{yogatama2017generative,zhang2017generalized}, and Transformer \cite{guo2019gaussian,dong2019unified}.  Scene/caption classification aims to classify texts in one image background or foreground. Roy et al. \cite{roy2017temporal} proposed temporal integration for word-wise caption and scene text identification. Ghosh et al. \cite{ghosh2019identifying} proposed identifying the presence of graphical text in scene images using CNN.

\begin{figure}[htb]

\begin{minipage}[b]{.48\linewidth}
  \centering
  \centerline{\includegraphics[width=4.0cm]{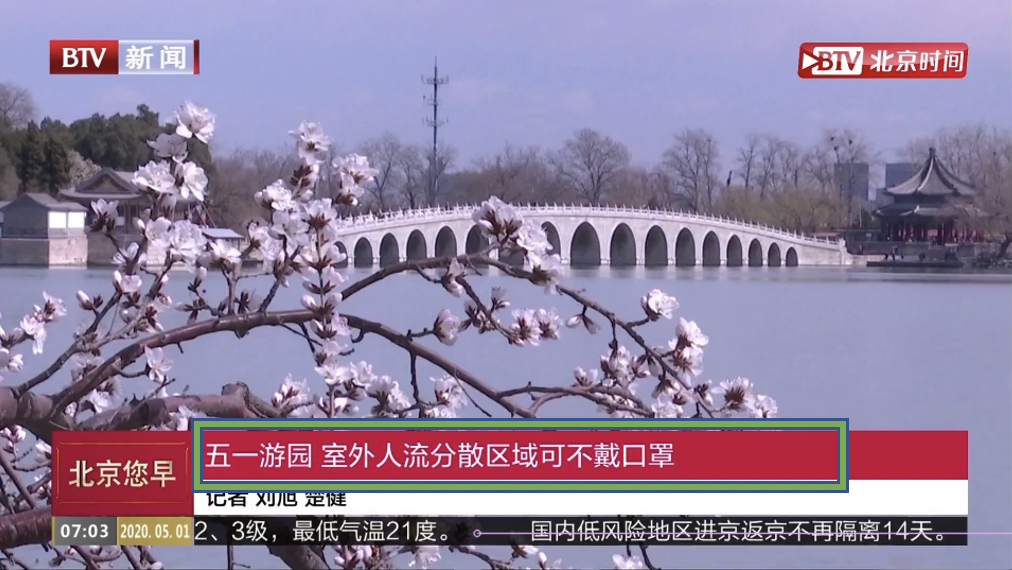}}
  \centerline{(a) an example of caption}\medskip
\end{minipage}
\hfill
\begin{minipage}[b]{0.48\linewidth}
  \centering
  \centerline{\includegraphics[width=4.0cm]{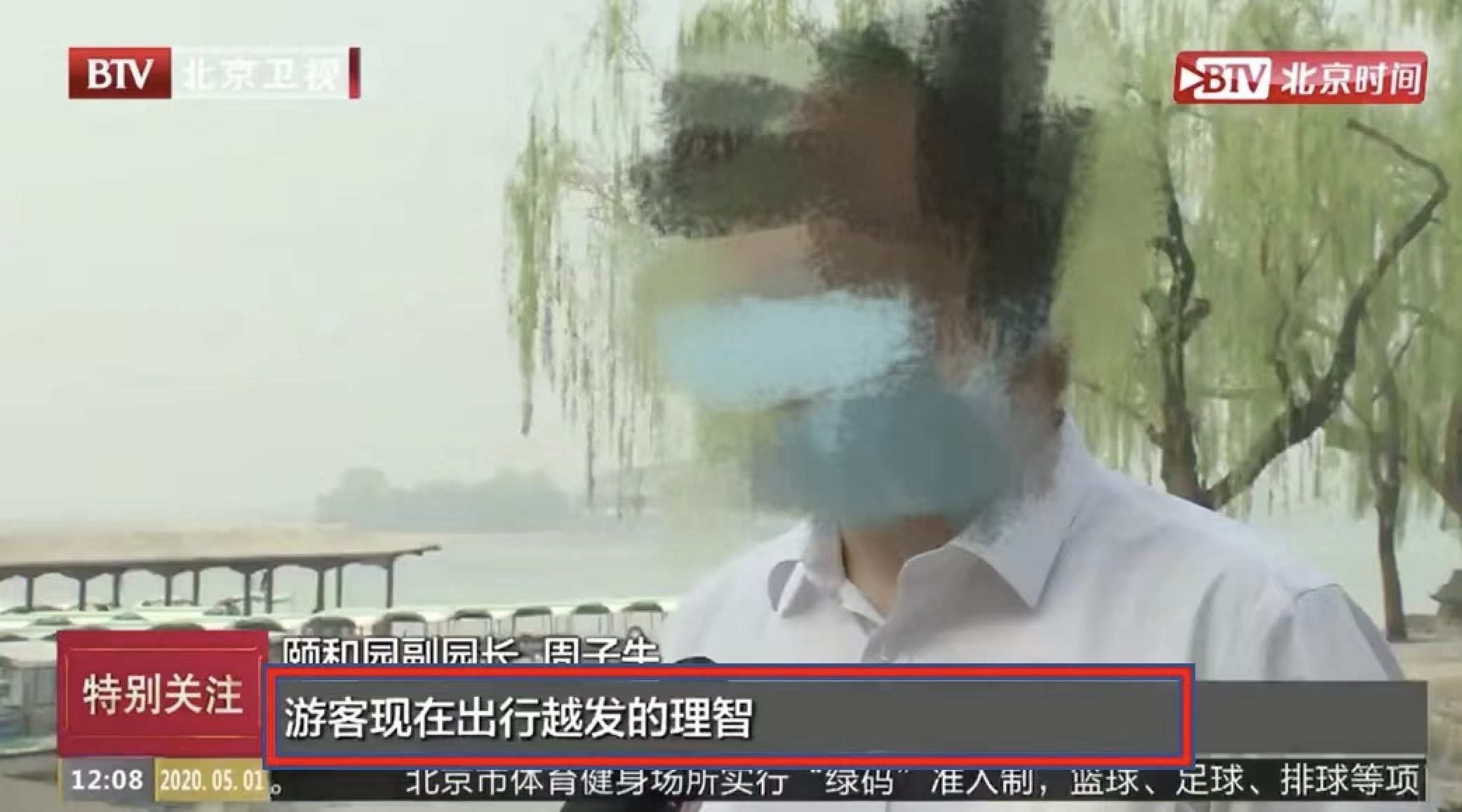}}
  \centerline{(b) an example of subtitle}\medskip
\end{minipage}
\caption{Examples of caption and subtitle on  video frames. Text in green box on the left belongs to caption, while the red one on the right is a subtitle text.}
\label{fig:caption_subtitle}
\end{figure}

Unfortunately, these approaches have important shortcomings which fail to get satisfying performance. Text classification based methods only take text as their input, ignoring other important modalities with critical discrimination information for some classes, e.g. caption and subtitle. Fig.\ref{fig:caption_subtitle}(a) and Fig.\ref{fig:caption_subtitle}(b) show an example of caption and  subtitle on one video frame, described as ``it 
is not necessary to wear mask in an outdoor area with dispersing crowd during Chinese May Day" and ``now tourists travel rationally", respectively. In this case, it is difficult to distinguish from one another based only on  text information, thus text classification methods  usually collapse with weak discrimination. Further, Fig.\ref{fig:caption_subtitle} shows that these two texts  appear in various colors and position, inspiring us to exploit multi-modality of visual and coordinate information. Considering scene/caption classification, it is not a real classification task. It only focuses on identifying a text background or foreground, thus this coarse-grained task does not classify texts' categories.  Consequently, a number of useless texts, such as rolling text in news videos, have no effect or even side effect on downstream video tasks.



In order to effectively assign the label of each text in videos, in this paper, we propose a novel \textbf{C}ontrastive  \textbf{G}raph \textbf{M}ultimodal  \textbf{M}odel, called CGMM, to leverage multimodal information in a harmonious manner. A specific module called CorrelationNet is established to explicitly extract layout information to enhance feature representation, meanwhile, contrastive learning is employed to learn the general features from plentiful unlabeled samples. Due to the lacking of public datasets related to our task, we construct a new large-scale dataset, called \emph{TI-News}, from the news domain.

\section{Methodology}
\label{sec:method}

\begin{figure}[h]
  \centering
  \includegraphics[width=0.8\linewidth]{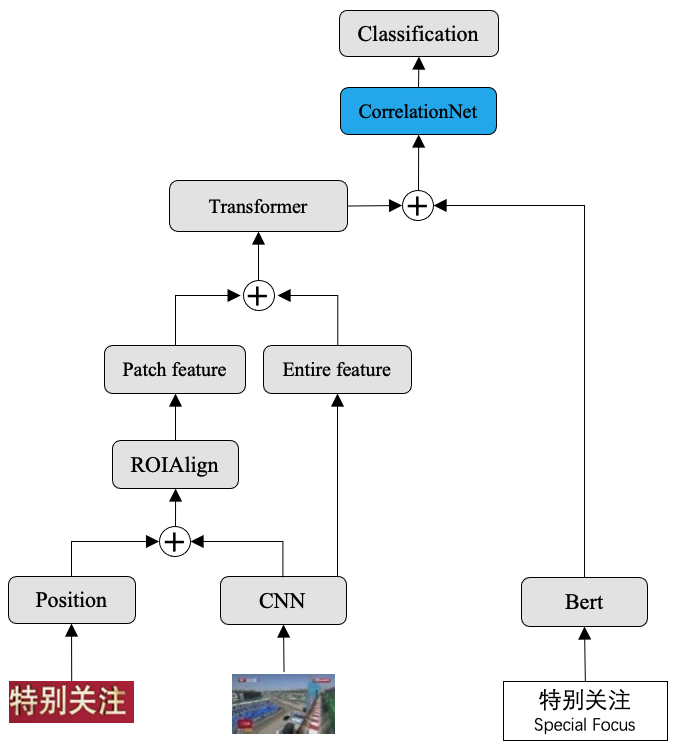}
  \caption{The schematic diagram of our CGMM algorithm. \textcircled{+} represents the operation of concatenation. Examples of texts in this diagram come from one news program.}
  \label{FIG: algorithm}
  \vspace{-4ex}
\end{figure}

The schematic diagram of our CGMM algorithm is shown in Fig.\ref{FIG: algorithm}. The architecture of CGMM consists of four parts. Firstly, multimodal features of vision, position and text are extracted, where the corresponding backbone is pre-trained using contrastive learning on unlabeled videos. Then, features of different modalities are fused and the multimodal representations of text boxes are obtained. Furthermore, we develop a new module called CorrelationNet to aggregate the spatial information among neighboring texts. Finally, the classification head is used to tag the text so as to determine its category.

\subsection{Multimodal Feature Extraction}

As mentioned in section \ref{sec:intro},  multimodal signals can provide abundant discrimination information so that we need to extract the multimodal representation of the texts. In this paper, we mainly extract visual, textual and coordinate features.

For visual modality, in contrast to conventional approaches which usually used classical VGG \cite{simonyan2014very} or ResNet-based \cite{he2016deep} network, we construct a shallow neural network as the backbone. In fact, from Figure \ref{fig:caption_subtitle}, we observe that texts differ in  low-level features of colors, fonts and sizes, thus above deeper networks  are abandoned as high-level semantic information is extracted which is not our need. Correspondingly, BERT \cite{devlin2018bert} is  used as the backbone of textual extraction. Upper-left and bottom-right coordinates represent the position of text in the frame.

\subsection{Multimodal Feature Fusion}

The fusion of multimodal features is a critical step to get the multimodal representation of video text. 
The process of visual and positional modalities is shown as bellow: 

\begin{equation}
f_{vis} = transformer(ROIAlign(f_{fra}, f_{pos}), f_{fra})
\end{equation}
where $f_{fra}$ and $f_{pos}$ as raw visual and positional features respectively. Firstly, ROIAlign \cite{he2017mask} is utilized to extract patch feature using  $f_{pos}$ and $f_{fra}$ as its inputs. Then, we use a transformer \cite{vaswani2017attention}   to  learn implicit relation between a patch box and its corresponding full frame. Finally, the visual feature  $f_{vis}$ and textual feature $f_{text}$ are  simply concatenated.


\subsection{Contrastive Learning}
\begin{figure*}[h]
  \centering
  \includegraphics[width=0.8\linewidth]{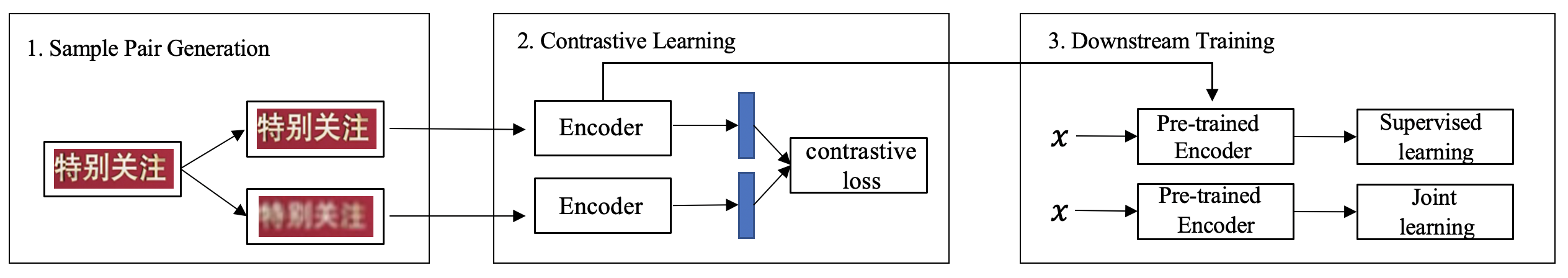}
  \caption{Overview of the contrastive learning framework.}
  \label{FIG: contrastive}
  \vspace{-4ex}
\end{figure*}

Due to limited labeled data, we elaborately utilize a contrastive learning \cite{le2020contrastive} strategy to fully explore inherent connections between samples using plentiful unlabeled videos. 
Fig.\ref{FIG: contrastive} illustrates the overview of the contrastive learning framework. 
It learns a representation which  embed positive input pairs nearby, while pushing negative pairs far apart.

Given a batch of $N$ samples $x_1,x_2,\dots,x_N$, the positive sample $\hat{x_{i}}$ of an anchor sample $x_i$ is established by disturbing its coordinate, changing image color, or replacing text with synonyms in lexical combinations. Thus a positive input pair is combined as $<x_{i},\hat{x_{i}}>$. The negative pair is formed as $<x_{i},x_{j}>$, where $j \in \{1,\dots,N\}, j \neq i$. 
Positive and negative sample pairs are fed to contrastive learning frameworks.

Based on the trained backbone of unimodal by contrastive learning, we attempt to train video text classification model using the labeled videos by two strategies: fine-tuning and joint learning.  Fine-tuning only uses traditional supervised learning to deal with our task.
Joint learning  combines the self-supervised contrastive learning and original supervised learning. The joint loss is denoted as follows:


\begin{equation}
L_{joint}= \alpha L_{cont} + (1-\alpha)L_{sup}
\end{equation}
\noindent where $L_{cont}$ is the contrastive loss defined in \cite{le2020contrastive} , and $L_{sup}$ is the classical cross entropy loss. $\alpha$ is the weight to balance these two losses.  

\subsection{CorrelationNet}
Edited texts from news videos, such as texts of caption and subtitle,  usually have regular layout. Figure \ref{fig:caption_subtitle} shows examples  of  caption  and  subtitle  on  some video  frames. It is observed that the
text of caption appears above that of person information, displayed in Fig.\ref{fig:caption_subtitle}(a), while the text of person information appears above that of subtitle, shown in Fig.\ref{fig:caption_subtitle}(b). This phenomenon inspires us to utilize the structural information to reinforce the feature representation. Motivated by Graph Neural Network \cite{scarselli2008graph}, CorrelationNet is proposed to model the relations among adjacent texts. 

\begin{figure}[h]
  \centering
  \includegraphics[width=0.65\linewidth]{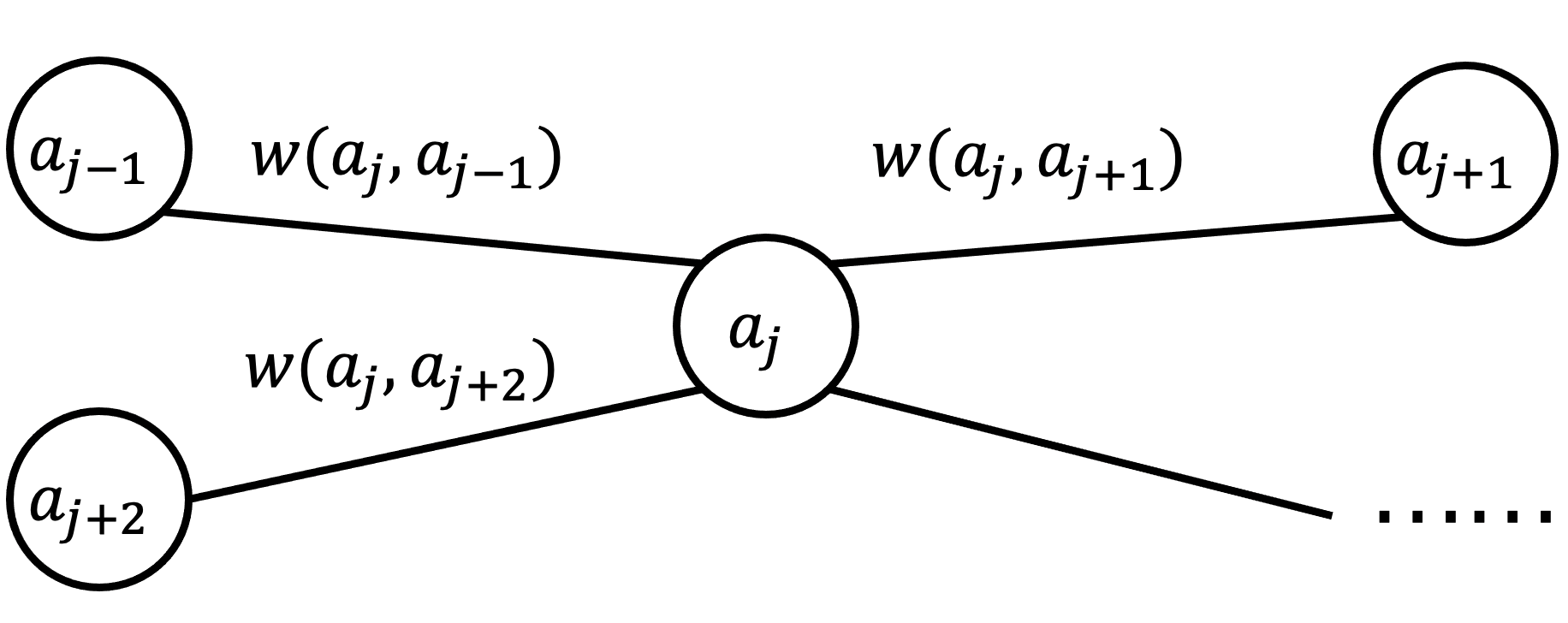}
  \caption{The graph for learning aggregated feature.}
  \label{FIG:aggregated feature}
\end{figure}

For a certain text box $a_{j}$, consider its adjacent neighbors $a_{1},\dots,a_{N}$. The total $N$ text boxes are used to generate the aggregated feature of $f^{(b)}(a_{j})$. 
Features of $N$ neighboring text boxes are weighted by a attention mechanism and then summed to get the fused feature. The graph constructed by the $N$ text boxes is illustrated in Fig.\ref{FIG:aggregated feature}.

$w(a_{j},a_{k})$ is denoted as the weight between two text boxes $a_{j}$ and $a_{k}$. Concretely, features are transformed by a fully connected network and then do subtraction to get a representation measuring the difference. Finally, a multilayer perceptron (MLP) is applied to get the weight.

After calculating all the weights between $a_{j}$ and its neighboring texts, the aggregated feature is obtained:

\begin{equation}
f^{(b)}(a_{j})=\sum_{i\in A(a_{j})}w(a_{j},a_{i})f^{(1)}(a_{i})
\end{equation}

\noindent where $A(a_{j})=\{1, 2, \dots, N\}$, and $f^{(1)}(a_{i})$ is original feature of text box $a_{i}$.

By using weighted aggregation strategy, CorrelationNet can learn the contribution of each modality by highlighting its importance. The final multimodal feature $a_{j}$, which is fed to classification head, is:

\begin{equation}
f^{(final)}(a_{j})=concat(f^{(b)}(a_{j}),f^{(1)}(a_{j}))
\end{equation}

\section{Experiments}

\subsection{Dataset}


There are no publicly available datasets with unified annotations for video text recognition and  classification. To  promote this new area, we construct a new well-defined  dataset of  news videos, called TI-News, which is dedicated especially to studying video text recognition and classification tasks jointly. 
Over 450000 text-box samples are annotated in TI-News, which are extracted from over 100 videos in 23 news programs. In training phase, 350000 samples come from 90 videos related to 8 programs. In testing phase, 100000 samples are selected for evaluation, from which 50000 samples are
collected from the same programs with the training dataset to construct TI-News standard dataset, while 50000 samples are collected from  extra 15 distinct videos to  construct TI-News generalization dataset for  generalization testing. 


We use a general OCR engine to locate and recognize all the texts in videos. Then, an annotation system for video text classification task is designed. Three representative categories are defined. They are Caption, Subtitle and Person information. In addition, class Others is used to denote the text boxes which not belong to above-mentioned categories. Besides, abundant kinds of pre-trained models involved in visual, audio and text features extraction are presented, which are pre-trained by millions of short videos using contrastive learning. Therefore, the pre-trained models have strong ability of feature representation.



\subsection{Training Details}

For visual modality, video frame is first resized to the shape of $384\times480\times3$. A simple model building with 3-layers CNN is used as feature extractor. Correspondingly, a 4-TransformerLayers BERT with embedding dimension of 768 is used to extract textual features.  
A stacked transformer block with 2 TransformerEncoderLayers (d\_model=256, nhead=8, dim\_feedforward=256) is used to implicitly learn the relation between patch box and  full image. Adam is chosen as the optimizer with a learning rate of 0.0002. 

\subsection{Evaluation Criteria}
We use standard precision, recall and F1 score to evaluate the performance of CGMM.  To exploit the effectiveness of several designed modules, ablation study is applied.  Experiments on TI-News standard set and  generalization set are conducted.

\subsection{Results on TI-News dataset}


Results on TI-News standard dataset and TI-News generalization dataset are shown in Table \ref{tab:experiment standard} and Table \ref{tab:experiment generlize} respectively. In Table \ref{tab:experiment standard}, the precision and recall of CGMM exceed 89.05\% and 92.34\% in standard set, with 90.67\% as its F1 score, which lead to a splendid performance of video-text classification task. 
Moreover, it is easy to find that we can’t get good results with single modality.  In Table \ref{tab:experiment generlize}, the precision and recall of CGMM exceed 84.33\% and 87.44\% in generalization set, with 85.86\% as its F1 score, which  shows that our method has good generalization. If only consider the visual modality, the results are shown to be an random distribution. By incorporating text modality, precision has increased a lot in the category of caption and subtitle. 
Furthermore, CorrelationNet can aggregate  features of neighboring text boxes, which gives great help to the classification model. Also, contrastive learning is effective to enhance the modality representations. 


\begin{table} \small
\centering
\begin{tabular}{l|l|l|l} 
\hline
\hline
Methods              & \multicolumn{3}{l}{Classification}  \\ 
\hline
                     & Precision & Recall & F1            \\ 
\hline
CGMM~w/o CV        & 76.14     & 86.65  & 81.05         \\ 
\hline
CGMM~w/o NLP              & 40.63     & 32.11  & 35.87         \\ 
\hline
CGMM~w/o POS      &  86.41    & 92.15   & 89.19        \\ 
\hline
CGMM~w/o CorrelationNet    &  87.50    & 91.44  & 89.43         \\ 
\hline
CGMM~w/o Contrasive learning        & 86.96    & 92.02   & 89.42         \\ 
\hline
CGMM proposed    & \textbf{89.05 }  & \textbf{92.34 } & \textbf{90.67}   \\
\hline
\hline
\end{tabular}
\caption{Classification results on  standard dataset. }
\label{tab:experiment standard}
\end{table}

\begin{table} \small
\centering
\begin{tabular}{l|l|l|l} 
\hline
\hline
Methods              & \multicolumn{3}{l}{Classification}  \\ 
\hline
                     & Precision & Recall & F1            \\ 
\hline
CGMM~w/o CV        & 72.10     & 82.05  & 76.76         \\ 
\hline
CGMM~w/o NLP              & 38.47     & 30.41  & 33.97         \\ 
\hline
CGMM~w/o POS      &  81.83    & \textbf{89.55}   & 85.52        \\ 
\hline
CGMM~w/o CorrelationNet    &  82.85    & 86.60  & 84.69         \\ 
\hline
CGMM~w/o Contrasive learning        & 82.34    & 87.92   & 85.04         \\ 
\hline
CGMM proposed    & \textbf{84.33 }  & 87.44 & \textbf{85.86}   \\
\hline
\hline
\end{tabular}
\caption{Classification results on   generalization dataset.}
\label{tab:experiment generlize}
\end{table}

\subsection{Ablation Study}
\noindent \textbf{Backbone} We study how the change of backbone  affects the performance of the proposed CGMM, as shown in Table \ref{tab:experiment 2}. In the proposed CGMM, a simple 3-layers CNN and 4-layers BERT are employed as the backbone of visual and textual modality, respectively. First, 3-layers CNN is replaced by a mobilenetV2 network, which leads to a drop of 7.5\% on F1 score against the proposed CGMM. 
Afterward, for the backbone of textual stream, we deepen BERT network from 4 layers to 8 layers. It is observed that 4 layers BERT is better. 

\begin{table} \footnotesize
  \centering
\begin{tabular}{l|l|l|l|l} 
\hline
\hline
Methods              & \multicolumn{4}{c}{Classification}  \\ 
\hline
                    & Backbone  & P & R & F1            \\ 
\hline
CGMM & CV(MobilenetV2)     & 80.13    & 86.34  & 83.12        \\ 
\hline
CGMM & NLP(8layers)            & 88.65     & 92.01  & 90.3         \\ 
\hline
CGMM proposed    &  --   & \textbf{89.05}    & \textbf{92.34} & \textbf{90.67}   \\
\hline
\hline
\end{tabular}
\caption{Ablation study of backbone on  standard dataset.}
\label{tab:experiment 2}
\end{table}

\noindent \textbf{Contrastive Learning} We empirically analyze the
effect of contrastive learning. Table \ref{tab:experiment 3} illustrates the  results from  three aspects: CV, NLP and position. We construct variants of  positive samples by disturbing its coordinates, changing image color, or 
replacing text with synonyms in lexical combinations.  The results indicate that contrastive learning involved in all  three aspects can achieve the best performance. 
 
\begin{table} \small
  \centering
\begin{tabular}{l|l|l|l|l} 
\hline
\hline
Methods              & \multicolumn{4}{c}{Classification}  \\ 
\hline
                    & Contrastive  & P & R & F1            \\ 
\hline
CGMM & CV     & 88.23     & 90.98  & 89.58        \\ 
\hline
CGMM & CV+POS            & 88.47    & 91.77  &  90.09      \\ 
\hline
CGMM proposed    &  --   & \textbf{89.05}    & \textbf{92.34} & \textbf{90.67}   \\
\hline
\hline
\end{tabular}
\caption{Ablation study of contrastive learning on  standard dataset.}
\label{tab:experiment 3}
\end{table}


\section{Conclusion}

The paper solves a new video text classification problem which aims to assign the
label of each text. To improve the discrimination ability, we propose a multimodal network, called CGMM, to fuse multimodal information of vision, text and position, into an unified framework. 
To explore layout information, a new module CorrelationNet is developed to use latent correlations among neighboring texts. Furthermore,  contrastive learning is used to strength the representation using  unlabeled videos. For news program application, a new dataset, TI-News, is established. The experiments on TI-News verify the effectiveness of our method.

\vfill\pagebreak

\bibliographystyle{IEEEbib}
\bibliography{icassp}

\end{document}